# Modeling Instantaneous Changes In Natural Scenes

**Vikram Dhillon**

**ABSTRACT**

This project aims to create 3d model of the natural world and model changes in it instantaneously. A framework for modeling instantaneous changes natural scenes in real time using Lagrangian Particle Framework and a fluid-particle grid approach is presented. This project is presented in the form of a proof-based system where we show that the design is very much possible but currently we only have selective scripts that accomplish the given job, a complete software however is still under work. This research can be divided into 3 distinct sections: the first one discusses a multi-camera rig that can measure ego-motion accurately up to 88%, how this device becomes the backbone of our framework, and some improvements devised to optimize a know framework for depth maps and 3d structure estimation from a single still image called make3d. The second part discusses the fluid-particle framework to model natural scenes, presents some algorithms that we are using to accomplish this task and we show how an application of our framework can extend make3d to model natural scenes in real time. This part of the research constructs a bridge between computer vision and computer graphics so that now ideas, answers and intuitions that arose in the domain of computer graphics can now be applied to computer vision and natural modeling. The final part of this research improves upon what might become the first general purpose vision system using deep belief architectures and provides another framework to improve the lower bound on training images for boosting by using a variation of Restricted Boltzmann machines (RBM). We also discuss other applications that might arise from our work in these areas.

# 1. INTRODUCTION

A lot of work has been done in the field of depth map estimation and motion modeling through 3d structure reconstruction mostly using stereo cameras and sometimes even use tedious setups such as LCD projectors which make it unreliable for practical use. Most of these algorithms also make assumptions about the environment but an interesting piece of work is presented in [1,2] where the authors present a framework called make3d capable of reconstructing a visually appealing 3d fly-throughs from a single still image. Make3d only assumes that made up of a number of small planes and it starts with breaking the image up into a number of tiny patches. According to [1,2] for each patch in the image, a Markov Random Field (MRF) is used to infer a set of "plane parameters" that capture both the 3-d location and 3-d orientation of the patch using supervised learning and after the parameters are learned we use them to construct the fly-throughs. Although visually appealing the mechanism only works on a single image and in addition to that its also computationally expensive. Due to that limitation we can't model the changes that occur in the subsequent frames after that one snapshot is taken. Modeling what happens in subsequent frames is a very difficult task because a vast amount of factors play a role in this for instance introduction of a new object, changing of the whole background and partial occlusions. In the following sections using a divide and conquer approach this bigger task of modeling is broken down into simple and more manageable tasks using the multi-camera rig. The organization of this paper is as follows: Section 2 discusses the multi-camera rig, section 3 presents the fluid-particle framework, section 4 discusses the general purpose vision system and some possible applications of our work and we conclude in section 5.

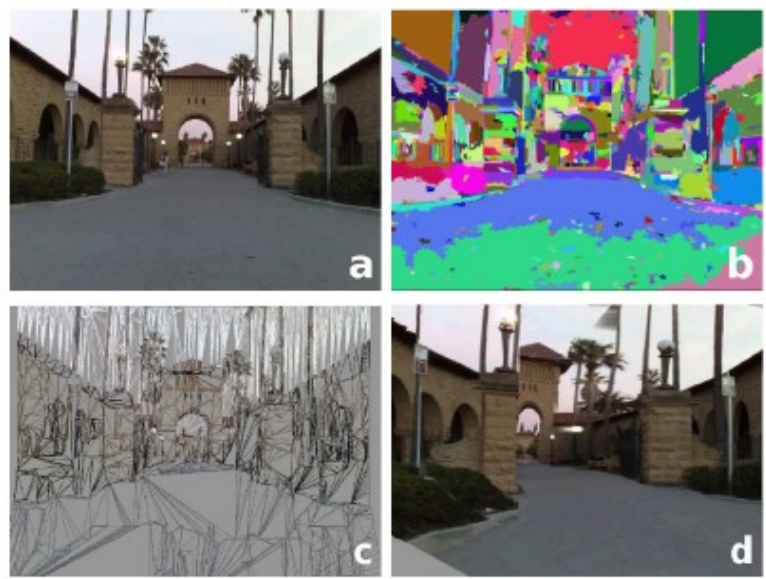

**Fig. 1. Make3d in action, taken from [2] (a) An original image. (b) Oversegmentation of the image to obtain "superpixels". (c) The 3-d model predicted by the algorithm. (d) A screenshot of the textured 3-d model.**

## 2. EGO-MOTION CALCULATION

Ego-motion estimation is tracking the motion of a camera and it has been an open research problem in computer vision for a long time. The main hurdle in this task is that when you start moving the camera around, the camera can't interpret if the motion being made is rotational or translational, this is also sometimes known as the "ambiguity problem". From [18] for a given point **P**, in the Cartesian coordinate system the velocity vector **V** is given by the following relationship

$$V = -T - \omega \times r. \qquad (1)$$

where $T$ is the translational vector, $\omega$ is the rotational component and r is the position of the vector in the coordinate system. The components of $T$ and $\omega$ are

$$T = (U, V, Z)^T \text{ and } \omega = (A, B, C)^T \qquad (2)$$

Now from (2) we can write (1) in its component form

$$\begin{aligned} X' &= -U - BZ + CY, \\ Y' &= -V - CX + AZ, \\ Z' &= -W - AY + BX. \end{aligned} \qquad (3)$$

where ' denotes differentiation with respect to time. Bruss and Horn formulate this problem mathematically and construct (3) which is now called the instantaneous model of optical flow. It provides a method for solving the translational and rotational ambiguity using least-squared regression but their main objective was to solve the problem of passive navigation. Nevertheless, their seminal work laid the background for later developments, later on Tsao *et. al* in [16] show a multi-camera model that is capable of solving the ambiguity problem with accuracy but it was very limited in the magnitude of translation and motion. Following [16] for a camera $C_k$ that is shifted from the origin the translation $t^k$ is given by

$$t^k = m_k[(\omega \times s_k) + T], \quad \omega^k = m_k \omega \qquad (4)$$

where $t_k$ is the translation and $\omega_k$ is the angular velocity of the camera $C_k$ placed at position $s_k$ with orientation $m_k$ and $T$ is the translation. Recently work done by Gupta *et. al* in

[20] presents a multi-camera rig that is capable of calculating ego-motion very efficiently and is applicable to wide range of motion by combining the two previously mentioned works and using a concept called polar correlation. For a given flow vector $[u^k, v^k]^T$ the optical flow from the instantaneous model is calculated by

$$u^1 = \frac{-\omega_y - T_x + xT_z}{Z} + \omega_x xy - \omega_y(x^2+1),$$
$$v^1 = \frac{-\omega_x - T_y + yT_z}{Z} + \omega_x(y^2+1) - \omega_y xy - \omega_z x. \tag{5}$$

Tsao *et. al* in [16] made a wonderful observation that given a camera motion if you have multiple cameras, the optical flows observed from the left and the right camera can solve the ambiguity problem. If the flows are the same in scale but opposite in direction, then the motion is translation otherwise its rotational. After calculating the optical flow, the translation is scaled where we cancel out all the rotational terms and are only left with translational terms as presented in [20]. We also use another concept presented by Gupta *et. al* called polar correlation basically states that opposing cameras have a common component of optical flow that can be canceled out to get the direction of the camera rig. Now that we have established a device capable of calculating ego-motion reliably we will see how this device becomes crucial in modeling instantaneous changes in the following section.

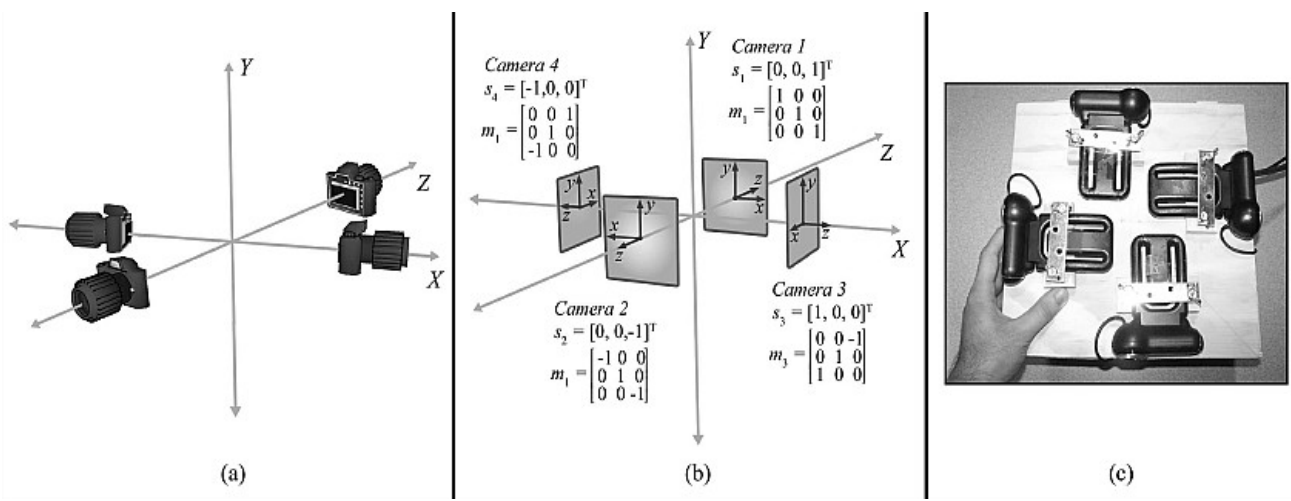

**Fig. 2. Multi-camera rig, taken from [20] (a) Diagram of the rig. (b) Position and orientation of the cameras. (c) The rig itself**

## 3. FLUID-PARTICLE FRAMEWORK

In this section we present the fluid-particle framework and then an extension is applied to make3d that helps achieve instantaneous modeling. This framework consists of a method to capture these changes, transfer them to the computer and modeling them in 3d. To do so, first we need a method to collect and represent those changes, make3d does an excellent job of capturing the depth information but only from a single frame, but more flexibility is needed for an extension to multiple frames. We also want to choose a material for modeling that deforms fast and model them accurately while holding the shape therefore a viscoelastic fluid is a good choice, the main advantage of doing so is to keep track of the changes that occurred in the video in the $t-1$ frame. Optical flow is the best way to see what changes occurred from frame $i \to i+1$ so we first start off making a displacement bridge (for the capturing stage) a construction that would help us efficiently and accurately capture the displacement vectors from optical flow.

***Lemma 1.1*** *Given two video frames, changes occurred can be captured by displacement vectors in Farnebäck's dense optical flow algorithm, as shown in [27]*

***Proof:*** *This follows from the very definition of optical flow.*

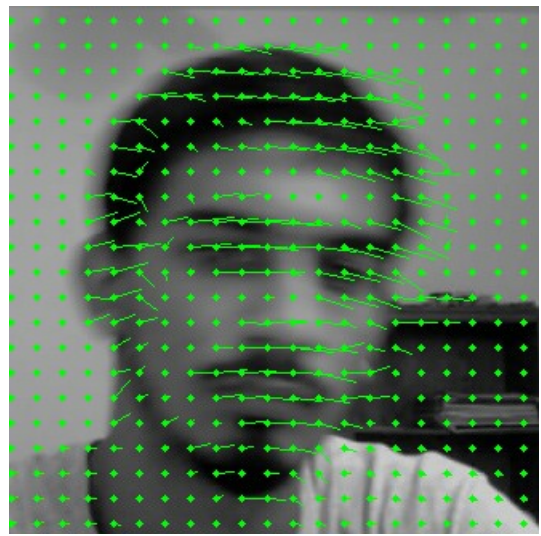

**Fig. 4. An example of Farnebäck's dense optical flow algorithm at work. Notice how the displacement vectors change direction to show the movement of the head.**

***Lemma 1.2*** *In the make3d framework, reconstruction should be possible using a different base than texture or color. Here by a different base we refer to the programmatic steps in the algorithm that reconstruct the 3d model in make3d.*

*Proof: Make3d must use some temporary variable(s) to store the predicted 3d model as shown in Fig. 1.（c）Following this reconstruction in the code it should be possible to modify the existing code to use a fluid for reconstruction rather than using colored texture. We will show later that a very accurate and detail descriptive model isn't needed for this job.*

This is a critical part in our remodeling, the fluid reconstruction here makes a foundation upon which our changes will be modeled. We split up the modeling stages in two fluid and particle stages. Starting off with fluid modeling we now need to transfer this displacement information, to do so we represent video motion or displacement in this case as particles that are being advected by the average general direction of the optical flow. We needed something that can respond to changes very fast, a solid texture reconstruction as used in make3d wouldn't change at all, the next best material that could be used was fluid, fluid can handle the changes efficiently but it won't model them fast, also one problem with fluid is that its moves with the average motion of its neighbor, so a simple change in one part of the model would cause the whole model to distort. Therefore we needed an approach that could avoid the distortion and still be capable of modeling the changes. A particle-based approach offers this kind of flexibility. Following [12] we use a frame-to-frame based optical flow field and using Lemma 1.2 we make another base of particles each one having some epsilon euclidean distance between them. Following [13] a Lagrangian particle-grid is then initialized over the optical flow field and particles are advected using the displacement vectors from Lemma 1.1, we captures the LCS (described below) in the motion flow.

**Lemma 1.3** *Given the advection of particles in a flow field Lagrangian Coherent Structures (LCS) can be found using a Lyapunov Exponent approach*

*Proof: Coherent Structures (CS) in a flow can be discovered using advection, these structures influence the movement of particles over time and divide them into distinct regions where all particles in a similar region have the same coherent behavior [9]*

Our particle system is a 4-grid approach and the layers help build each other. Structurally each layer is the same, each has a 2D picture of that was given to make3d and particles that were initialized in the layer 1 (following). The first layer's function is to get the advection of particles from the optical flow vectors. Layer two captures the LCS from layer 1 as shown in [9] we are using their idea here but in a different application as a part of the layer.  Layer 3 has an extra feature that other layers don't, its weight sensitive, the LCS once captured from the layer 2, this particles are assigned similar weight if they belong to the same LCS group and the advection as usual follows from layer 1. Layer four is what we constructed from Lemma 1.2 and is the top layer, the particles in this layer move with average motion of its neighbor which is directed by the third layer. The average motion and LCS ensure the correction of some errors that may have occurred in the translation of vectors and  bring smoothness to the advection.

***Lemma 1.4*** *Given a 3d model constructed by make3d, we can find/retrieve the depth of each pixel or patch in the model from the MRF*

***Proof:*** *One of the beauties of the make3d architecture is that even though the input is a single still image, the algorithm's MRF records the depth of each patch. To record these depths make3d must use a surface as a reference frame to which the depths are relative to. This function is analogous to surrounding the model with a cube and using a face of the cube as the reference frame to which the depths are relative. Then make3d from that surface models the depths in the fly-through. We can retrieve those depths and thus notice how far each pixel is from the reference frame.*

Lemma 1.4 offers a very helpful feature, once the depths are learned we don't need original picture used for make3d modeling or the 3d fluid model constructed for that matter, we can now perform all the modeling we want on this two dimensional image and if we need the 3d model again we just stack multiple modified images and using  the depth and get a new 3d model from that. This hold true for all conditions but when there's a new object introduced in the scene, that's a disastrous thought at first because it can potentially destroy all the layers but we'll tackle this situation in the next section. Since the natural scenes are dynamic, we need to take in account all small motion that occurs in the scene we capture and may potentially lead to miscalculation which will then be passed on to each layer. To solve this problem we use [12] to devise a maintainability framework that is applied to layer 1, the other layers won't be needing all the steps because they operate on previously constructed layers. [12] discuss 5 major steps that can accomplish this and since there are subtle differences in the application of maintainability we discuss them here again. Note that maintainability only needs to be applied to particles because they are the carriers of dynamic information:

- *Propagation*: The propagation of particles form adjacent frame into the current frame is done according to forward and reverse flow fields. To propagate particle $i$ from frame $t-1$ to $t$ the following flow fields are used $u(x,y,t-1), v(x,y,t-1)$:

$$x_i(t) = x_i(t-1) + u(x_i(t-1), y_i(t-1), t-1),$$
$$y_i(t) = y_i(t-1) + v(x_i(t-1), y_i(t-1), t-1). \qquad (6)$$

  This was an example of forward propagation, backward propagation would be done in a similar way replacing the $t-1$ with $t$.

- *Linking*: The particles in layer 1 are subject to constant motion, and to keep their shape intact globally, we use a constrained Delaunay triangulation which ensures a good directional distribution of links between particles, avoids long-range drift of particles and spreads the motion on average across the linked particles. This is done by simply linking a particle to its *N* neighbors.

- *Optimization:* This is the core construction of the particle-maintainability algorithm, it uses energy based modeling to reposition particles after their initial position has been described by the flow fields. The particle objective function that describes the energy of a particle $i$ in frame $t$ is:

$$E(i,t) = \sum_{k \in K_i(t)} E_{Data}^{[k]}(i,t) + \alpha \sum_{j \in L_i(t)} E_{Distort}(i,j,t). \qquad (7)$$

  where $K_i(t)$ is the set of active channels and $L_i(t)$ is the number of particles linked to the particle $i$ in frame $t$. In the optimization process, the data term takes in account how the appearance of a particle changes to cope with non-Lambertian reflectance and changes in scale. The data term suggests that a particle's appearance changes slowly over time but remains independent of motion smoothness. The second term called the distortion energy is used because the data term can't uniquely constrain particle positions and therefore the distortion term reveals the relative motion of particles. Distortion applied to a single frame and while used in combination with the data term assists the particles that have moved together to continue to move together and particles that have moved differently to continue doing so, that's very useful in the layer 3. Also the distortion term allows the global motion of the camera to be irregular.

- *Pruning:* After optimization the particles that have very high energy are removed because those particles have generally drifted beyond their range or encountered an occlusion.

- Addition: After pruning the algorithm adds new particles between particles where gaps have been left and this is done using scale maps.

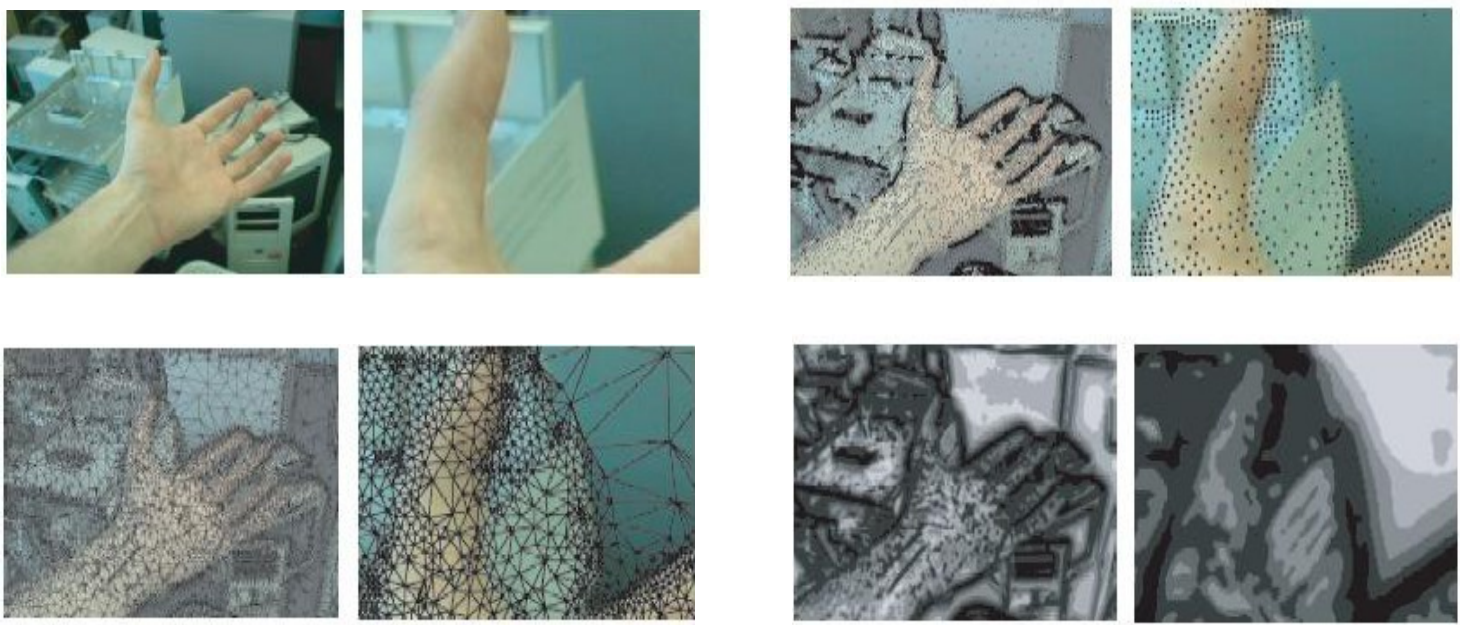

**Fig. 5. Particle maintainability algorithm at work. Taken from [12]. From left the first picture is the video frame, the second one initializes the particles, the third one shows the links, and the last one has the scale maps.**

.

So that describes the maintainability in particles, now we will shift our focus towards new object introduction, but first we need a mechanism to tell us whether a new object has been introduced or not. We introduce a new construction, following [25] we use the gist image representation for two basic purposes: detect if a new object has been introduced in the scene, and if the entire background has changed, both of which can malfunction the entire system. [15] thoroughly describes a concept called contextual priming which is basically using the context of an image to make inferences about the image itself. The main contribution of [15] is the gist image representation (GIR) and we use it in a different way, one not discussed in [15]. We use GIR as a binary classifier which makes the two tasks mentioned above easier, for the first task we just classify whether a new object is there or not, and for the second task we take in account the ego-motion and see if there are any changes across the y-axis describing a possible translation. From that , the gist will get a "hit" every time there is a new object in the scene. The incorporation of this new object will be different in particles and fluids therefore we take two separate approaches to them. On the fluid side, to capture a new object as the gist informs a new object has been introduced, we use a Fourier shape descriptor to get the shape of the object and then using a detail preserving fluid reconstruction we construct a new fluid shape from that. Following [23] a binary mask is constructed for motion detection, which basically assigns 0's to all the objects that are static, and 1's to all the objects that are moving. This in combination with the gist once put together creates a binary image, from that we use a background tracing algorithm described in [23] and the Fourier shape descriptors extract the shape from the object. Now once we have this shape, optimization and addition steps of the particle maintainability will take care of new object introduction on the particle side.

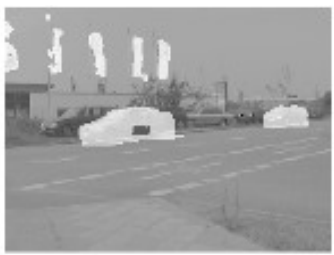 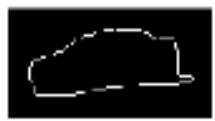

**Fig. 6. Fourier descriptors at work. Taken from [23]. The first picture describes the motion mask and the second one describes the shape extracted by the FD's.**

Each boundary pixel k is represented by the pair $(x(k), y(k))$ of its coordinates. Given an N – point boundary the complex numbers are defined by:

$$z(k) = x(k) + j \cdot y(k), \quad k = 0, 1, \ldots, N-1 \quad (8)$$

The x–axis is treated as the real axis and the y–axis as the imaginary axis of the sequence of complex numbers $z(k)$. The discrete Fourier transform of $z(k)$ is given by:

$$a(n) = \frac{1}{N} \sum_{k=0}^{N-1} z(k) e^{(-j2\pi nk/N)}, \quad n = 0, 1, \ldots N-1 \quad (9)$$

The complex coefficients a(n) are called the Fourier descriptors (FDs) of the corresponding boundary. They contain the same information about the object shape as the initial coefficients z(k). [23] goes on the specify how $a(n)$ must be scaled to perform object classification but we use the shape obtained here in a different way therefore the scaling isn't needed also the output of this system is fed into another as input, to construct the shape. The detail preserving fluid control method is presented in [21] where Smoothed Particle Hydrodynamics is used to solve the Navier-Stokes equations. Then the control particles are fed the shape of the object that we got from the FD's and the fluid simulation generates the 3d object for the fluid maintainability. An overview of the approach take in [21] is presented below. Please refer to [21] for further discussion on this topic.

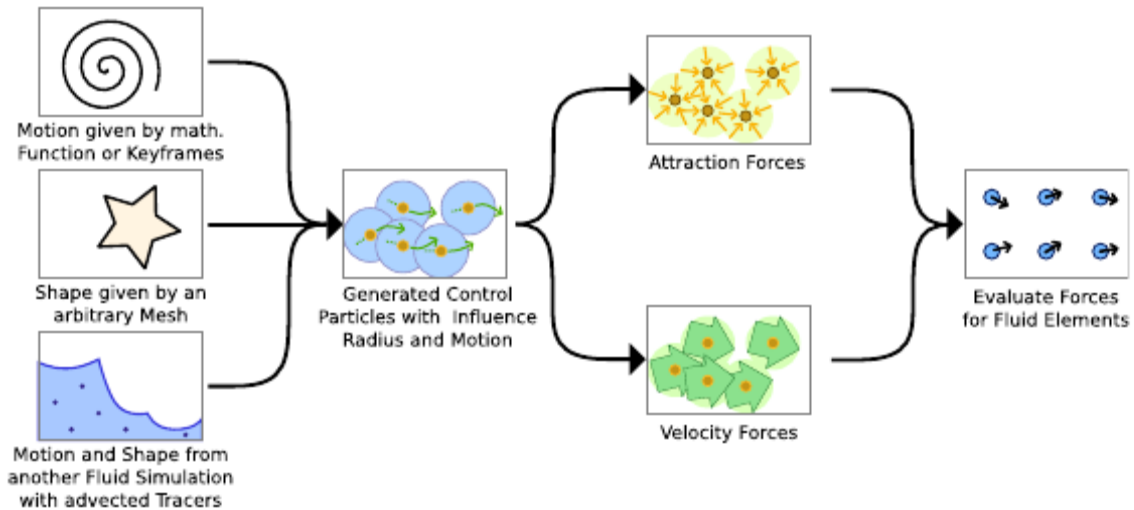

Fig. 7. Detail-preserving fluid control algorithm. Taken from [21]. These steps with slight modification fit our needs to generate the 3d space.

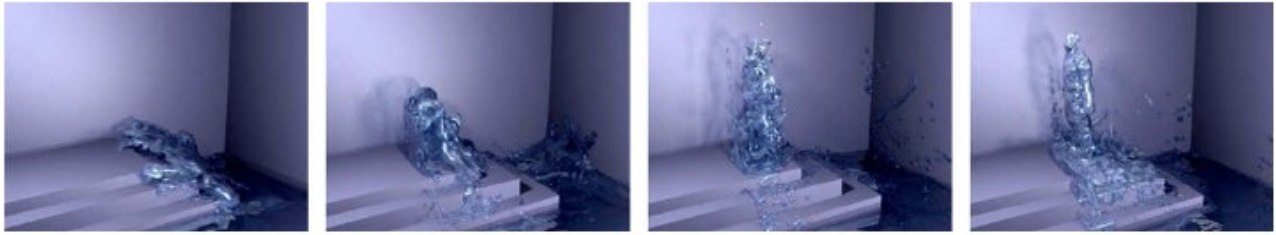

**Fig. 8. A sequence showing the result of the fluid control algorithm. Taken from [ 21]. It shows the water going up the stairs and forming the 3-d shape of a human figure as provided to the algorithm.**

One of the critical discoveries made by Thürey, Keiser *et. al.* [21] is that directly enforcing control from the particles onto the fluid can lead to noticeable distortions of the velocity field, which is noticeable as an increased viscosity. The artificial viscosity is avoided by decomposing the velocity field into coarse and fine scale components and the control forces are only applied to the low-frequency part. High-frequency components are largely unaffected, thus small-scale detail and turbulence are significantly better preserved. This will serve to be crucial when we perform the fluid deformation. All of the above holds true if we take one assumption, and that is the background remains the same, even if there's motion, the whole background must not change, now we will tackle a situation in which the background can change.

***Lemma 1.5*** *A change in ego-motion graph suggests the camera has been moved.*

***Proof:*** *This is evident from the definition of ego-motion, if the camera moves (translation or rotation), the ego-motion graph will plot its path*

Here's by using Lemma 1.5 we track whole background change, now that we have a device that can track ego-motion we can tell whether an entire background change has occurred, because if it did the ego-motion graph will show changes across the x-axis. We do account for slight movements made by the user while handling the camera as they don't show up on the ego-motion graph but a clear movement to the left on the other had will be recorded. So from the direction of the ego-motion, we take the first frame of the new scenery, send it to make3d and get a new 3d scene from it, then using the ego-motion we put it in its relative location and start tracking changes in both scenes. Now we move on the the task on transferring the changes that the particles captured on to the the fluid, the instantaneous changes that occur in the natural scenes will be captured by the particles but say at some instant in time we want to see the depth map again, before we discussed the concept of depth learning in Lemma 1.4 which we will use here. So when a pause button is pressed, we freeze the state of layer 4 and transfer the positions and orientation of particles onto a different space, which contains a surface as shows in Fig. 9., we stack up layer 4 to get a volumetric representation in this space and then start pouring in a viscoelastic fluid.

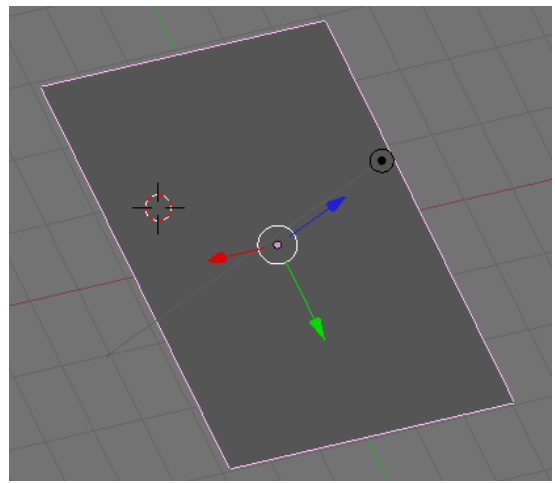

**Fig. 9. Planar space representation from Blender**

The fluid hence easily molds into the shape that the particles had and then we take a cross-section out of that, so now we have a fluid representation of what the particles looked like at the instant the freeze button was pressed. Agostino *et. al* in [13] presents the precise mathematical formulation of the deformation as a function of a force field acting upon it, please refer to [13] section 2 for a further discussion. Thus we present the fluid-particle grid framework capable of modeling instantaneous changes in natural scenes. Note that all the changes that occur are done in a 2d layer so as was stated in Lemma 1.2 some small errors in the construction won't cause a problem after the depths are learned accurately. See Appendix A. for a highlight of the process.

**4. DEEP BELIEF NETS AND A GENERAL PURPOSE VISION SYSTEM**

In this section we will talk about a recent development on neural nets called deep belief nets and how it relates to our task here. Hinton *et. al* in [8] presented a modification on Boltzmann machines with no lateral interactions which were called Restricted Boltzmann Machines (RBM) which can be trained efficiently using back propagation. He later on he goes on to state that this can be done in layers as well and also that every time we add a layer, if its done right, we will get better results from the training data. This multi-layer model of the RBM's is called a deep belief net and is mostly used for reconstruction as of now, DBN's are very powerful because of the fact that they can learn multiple layer generative models of data efficiently and accurately, also a higher level representation of the data can be built from unlabeled data which is much easier to acquire. In a DBN we train the first layer to reconstruct the sensory input, in our case we will use numbers, each of proceeding layers captures higher-level correlation between the activities of hidden features in the layer below and as we go higher and higher we expect better generative models of the interaction among the data. The top two layers however have symmetric connections among them and form an associative memory, and after the unsupervised training is done, a back propagation algorithm is used to fine-tune the weights on each connection going down the layers to make better reconstructions. Other procedures such as maximum likelihood learning are discussed in [10]. So taking on the task of reconstruction, we first discuss another framework to improve the lower bound on training data sets and assists in pre-processing so we can get better detection results, afterwards we will discuss how to apply this alongside GIR to get what might become the first general purpose vision system. To start off with the reconstruction framework, we first have to discuss a device called an auto-encoder, an auto-encoder learns a compressed representation (encoding) for a set of data and later on uncompresses it into its dimensions, most commonly used for dimensionality reduction. Auto-encoders have the structure of a basic neural net, they use three or more layers:

- An input layer. For example, in a face recognition task, the neurons in the input layer could map to pixels in the photograph.
- A number of considerably smaller hidden layers, which will form the encoding.
- An output layer, where each neuron has the same meaning as in the input layer.

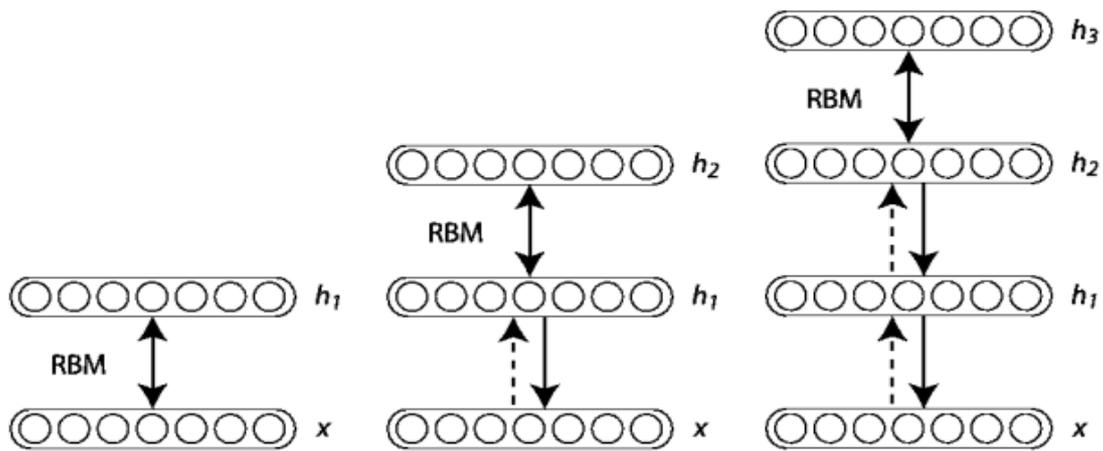

**Fig. 10. A schematic diagram of a RBM.** $h$ **here represents the hidden layers.**

We now construct a variation of the learning module CRBM which we call a hybrid net that works as follows, a gated CRBM is taken into use following [10] so the framework can also be applied to videos and not only images. We will separately discuss the two cases, in videos we intercept those red interactions on the lower layers with an auto-encoder and say for example we wanted to remove

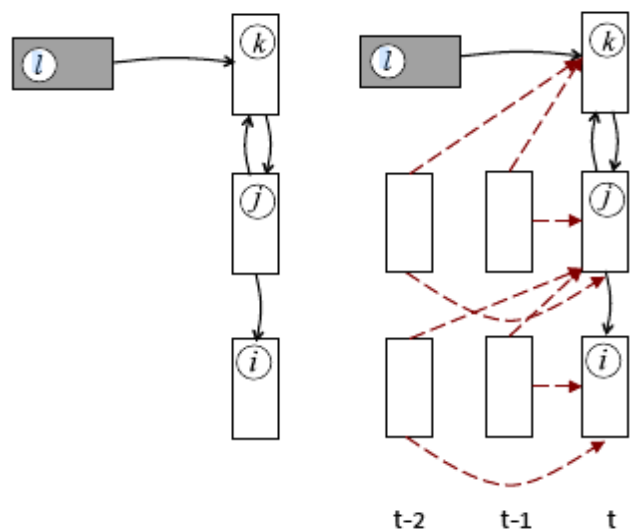

**Fig. 11. A schematic diagram of a CRMB. Taken from [10]. The red markings show the path taken by the message as it propagates from one to the next hidden layer**

noise, we would use a variation of auto-encoders called stacked denoising auto-encoder which is more robust to removing noise, and before the message reaches the next hidden unit, its cleared of the noise it had before, so now the reconstructions can be better and the overall generative model will turn out better. We also propose adding some decision units, that first evaluate the need to apply a pre-processing algorithm, continuing on our example of noise [26] presents an algorithm that can detect the presence of salt and pepper noise, our decision units will apply the detection stage of the algorithm and see if its necessary to even apply denoising. In doing so, the decision units help in reducing the time and number of algorithms that would otherwise have been used. For simple images this reconstruction framework will only use the time $t$ and by going through the entire hybrid net, we hope to get better reconstructions that the data itself which will dramatically increase the detection rates for other tasks. This was a big problem for [25], they have a very impressive system for detecting a multitude of objects using contextual priming but their only problem seemed to be that once a lot of small detectors were combined to make a big one, they would drop in the rate of detections. Now use a hybrid net we can solve that problem, the detections would be much better after the reconstructions, not only that this procedure can be extended to include other common processes for example inpainting but we won't discuss that here. Another improvement that we would like to bring to notice is that we can use back propagation here again to improve the results that we get after integrating our hybrid net with the system proposed in [25]. Another area where the hybrid net would be very useful is novelty assessment, as shows in [3] auto-encoders work nicely for this task but a hybrid net would work even better because of its very structure, the multiple interactions and the decision units. It would be easier to asses the distortion as when the normal data passes through the net it won't have much distortion but when a system encounters something completely new, the decision units at that time can measure the distortion and if combined with another supervised learning algorithm that can choose which action to take given some epsilon distortion it can automatically handle the "abnormality" that arise.

## 5. DISCUSSION

In this paper we have presented a device capable of capturing ego-motion that assists in detecting complete background changes for the fluid modeling, a fluid-particle framework capable of capturing instantaneous changes in the outside world and a deep belief net based framework for pre-processing of images. The fluid-particle framework leaves a lot of room for advancements as once a bridge between machine learning and computer animation has been made we can now use any technique that was present in computer animation to more accurately model a variety of phenomenon for example modeling random and unpredictable phenomenon using the same techniques as are used for modeling explosions. Future efforts consist of packaging the scripts we use now for the fluid-particle framework and the hybrid net into a complete software, and exploring other possibilities that can be modeled using the frameworks.

# 6. APPENDIX A.

Here we put together the whole fluid-particle framework and the functions it uses to accomplish the job, notice how each function is dependent on the one before it for input.

```
INPUT: raw video sequence and ego-motion
ALGORITHM:
    frame = { }, displacement = { }
    optical_flow (frame, displacement)           // calculate the optical flow
    displacement_bridge (optical_flow)           // transfers the optical flow
    initialize_particle_grids ( )                // initialize the 4 layers
    new_object_initialization ( )                // start the GIR and prepare for a hit (positive detection)
    make3d_fluid_reconstruction (frame 0)        // make the initial fluid model from the frame
    make3d_frame_reconstruction (frame 0)        // make a 3d model from the given frame
    make3d_particle_reconstruction (frame 0)     // make a 3d model using particles
    depth_splicing (make3d_frame_reconstruction) // learn depths
    // advect particles using optical flow to find LCS for the removable layer
    LCS_particle_advection (optical_flow, make3d_particle_reconstruction)
    LCS_capture  (LCS_particle_advection)        // capture LCS
    LCS_relocate  (LCS_capture)                  // relocate particles using the capturing mechanism
    particle_displacement (LCS_relocate)         //particles relocate by the average motion of LCS layer
    freeze_particles (particle_displacement)     // mechanism to freeze the particles position
    planar_transformation(freeze_particles)      // transfer the position of particles into a new space
    fluid_deformation (freeze_particles)         // follow the deformation process
    // prepare to incorporate new background
    background_change (ego-motion, make3d_frame_reconstruction)
```

## 7. APPENDIX B.

Make3d utilizes a lot of memory and structurally it has been written in such a way that optimizing parts of it is easy, following [24] we focus on a over-segmentation method that reduces accuracy by about 3% but gives results much faster. The PathFinder algorithm, as its called, is about 30-40 times faster [24] than the algorithm used for over-segmentation in make3d. Their approach is similar to superpixel lattice dynamic programming and they show that using dynamic programming you can achieve real-time results which are computationally cheap. They use a concept called strongest vertical paths to determine "perceptually similar" regions and the horizontal and vertical strongest components are used to determine a boundary for the superpixels.

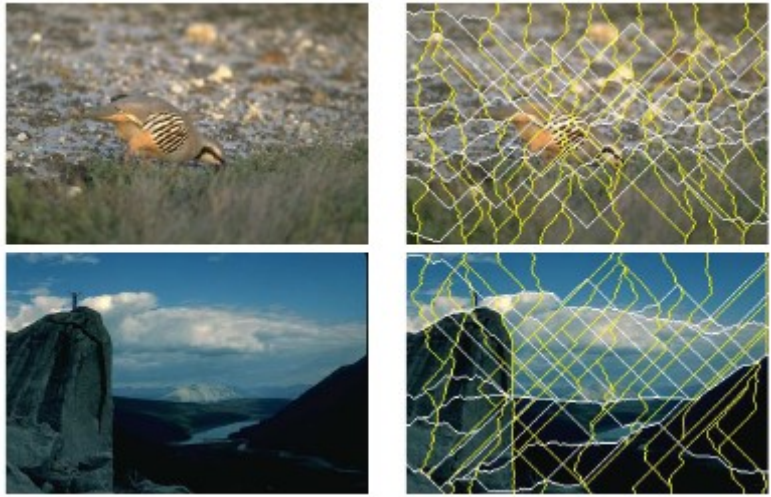

**Fig. 12. PathFinder at work. Taken from [24]. The yellow lines represent the boundaries for over-segmentation**

# 8. REFERENCES

[1] A. Saxena, S.H. Chung, and A. Y. Ng. 3-d depth reconstruction from a single still image. In *IJCAI*, 2007.

[2] A. Saxena, M. Sun, and A. Y. Ng. Make3d: learning 3d scene structure from a single still image. In *CVPR*, 2007

[3] B. B. Thompson *et. al.* Implicit Learning in Autoencoder Novelty Assessment. Proc. International Joint Conference on Neural Networks, Honolulu, pp. 2878-2883, May, 2002.

[4] T. Pock, M. Urschler, C. Zach, R. Beichel, and H. Bischof. A duality based algorithm for TV-L1 optical-flow image registration. In 10th International Conference on Medical Image Computing and Computer Assisted Intervention, pages 511–518, Brisbane, Australia, 2007.

[5] I. Sutskever , G. E. Hinton. Learning multilevel distributed representations for high dimensional sequences (Technical Report UTMLTR 2006-003). Dept. of Computer Science, University of Toronto.

[6] G.E. Hinton. Learning multiple layers of representation. Trends Cogn. Sci. 11, 428–434. 2007

[7] A. Grubb, J. A. Bagnell. Boosted Backpropagation Learning for Training Deep Modular Networks. Proc. International Conference on Machine Learning, Haifa, Isreal, 2010.

[8] R. Salakhutdinov and G. Hinton. Deep Boltzmann machines. In Proceedings of the International Conference on Artificial Intelligence and Statistics, volume 5, pages 448–455, 2009.

[9] S. Ali and M. Shah, A Lagrangian Particle Dynamics Approach for Crowd Flow Segmentation and Stability Analysis, IEEE CVPR, 2007.

[10] G. Taylor and G. E. Hinton. Factored Conditional Restricted Boltzmann Machines for Modeling Motion Style. In ICML, 2009.

[11] Erhan, D., Manzagol, P.-A., Bengio, Y., Bengio, S. & Vincent, P. (2009). The difficulty of training deep architectures and the effect of unsupervised pre-training. AI & Stat.'2009.

[12] Sand, P., Teller, S.: Particle Video: Long-Range Motion Estimation using Point Trajectories. Proc. Int'l Conf. Computer Vision and Pattern Recognition (2006) 2195-2202

[13] E. D. Agostino, F. Maes *et. al.* A viscous fluid model for multimodal non-rigid image registration using mutual information. Medical Image Analysis 7 (2003) 565-575.

[14] Klaas, M., Lang, D., & de Freitas, N. (2005). Fast maximum a posteriori inference in Monte Carlo state spaces. Artificial Intelligence and Statistics.

[15] A. Torralba, K. P. Murphy, W. T. Freeman, and M. A. Rubin. Context-based vision system for place and object recognition. In Proc. ICCV, 2003.

[16] A-T. Tsao, Y-P. Hung, C-S. Fuh, and Y-S. Chen. Ego-motion estimation using optical flow fields ob- served from multiple cameras. In IEEE Conference on Computer Vision and Pattern Recognition, San Juan (Porto Rico), pages 457-462. 1999.

[17] A. C. Poje, G. Haller, and I. Mezic´. The geometry and statistics of mixing in aperiodic flows. Phys. Fluids A 11, 2963–2968 (1999).

[18] A.R. Bruss and B.K.P. Horn. "Passive navigation." Comptr. Vision, Graphics, and Image Process. 21:3-20, 1983.

[19] Ramin Mehran, Brian Moore, Mubarak Shah. A Streakline Representation of Flow in Crowded Scenes, European Conference on Computer Vision (ECCV), Crete, Greece, 2010.

[20] Gupta, P. da Vitoria Lobo, N. Laviola, J.J. Markerless tracking using Polar Correlation of camera optical flow. Virtual Reality Conference (VR), 2010 IEEE.

[21] R. K eiser *et. al.* Detail-preserving fluid control. In 2006 Symposium on Computer Animation (2006), pp. 7–12.

[22] M. Muller *et. al.* Particle-based fluid simulation for interactive applications. In Proc. of the 2003 ACM SIGGRAPH/Eurographics Symposium on Computer Animation, 154–159. 2003.

[23] D. Toyh and T. Aach. Detection and recognition of moving objects using statistical motiondetection and Fourier descriptors. In ICAP 03. 2003.

[24] F. Drucker, J. MacCormick. Fast superpixels for video analysis. Motion and Video Computing, 2009. WMVC '09.

[25] K. Murphy, A *et. al*. Using the forest to see the trees: exploiting context for visual object detection and localization, NIPS. 2003.

[26] R.H. Chan, C. Ho and M. Nikolova. Salt-and-Pepper Noise Removal by Median-type Noise Detectors and Detail-preserving Regularization, to appear.

[27] Gunnar Farneb¨. Two-frame motion estimation based on a polynomial expansion, In Proc. of the 13th Scandinavian Conf. on Image Analysis, 2003, vol. 2749 of LNCS, pp. 363–370.